# Enhancing Deep Learning Models through Tensorization: A Comprehensive Survey and Framework

Manal Helal

mhelal@ieee.org
University of Hertfordshire, School of Physics,
Engineering and Computer Science, Hatfield, UK

**Abstract:**

The burgeoning growth of public domain data and the increasing complexity of deep learning model architectures have underscored the need for more efficient data representation and analysis techniques. This paper is motivated by the work of (Helal, 2023) and aims to present a comprehensive overview of tensorization. This transformative approach bridges the gap between the inherently multidimensional nature of data and the simplified 2-dimensional matrices commonly used in linear algebra-based machine learning algorithms. This paper explores the steps involved in tensorization, multidimensional data sources, various multiway analysis methods employed, and the benefits of these approaches. A small example of Blind Source Separation (BSS) is presented comparing 2-dimensional algorithms and a multiway algorithm in Python. Results indicate that multiway analysis is more expressive. Contrary to the intuition of the dimensionality curse, utilising multidimensional datasets in their native form and applying multiway analysis methods grounded in multilinear algebra reveal a profound capacity to capture intricate interrelationships among various dimensions while, surprisingly, reducing the number of model parameters and accelerating processing. A survey of the multi-away analysis methods and integration with various Deep Neural Networks models is presented using case studies in different application domains.

## 1. Introduction

The motivation to write the book (Helal, 2023) and summarise the tensorisation step in this paper is the increased size of the public domain data and the increased size of deep learning models' architecture. The data are intuitively multidimensional but simplified as a 2-dimension matrix form for simpler representation and application of linear algebra algorithms. Using the multidimensional dataset in their given form and applying multiway analysis methods using multilinear algebra provide better expressive models of the multiway interactions of the different dimensions, and surprisingly, lead to fewer parameters and faster processing, not the expected dimensionality curse from the increased dimensions. Various papers and projects following different standards present the current advances in multiway analysis. The main objective is to explain in order all the required theoretical background to understand these methods, provide a survey on the existing high dimensional datasets, tensorisation steps of 2-dimensional datasets, and the available methods for multiway analysis and compressed deep learning models.

This paper systematically navigates the essential theoretical foundations required to comprehend these methods, offering insight into the intricacies of high-dimensional datasets, tensorization procedures for transforming 2-dimensional data, and an extensive inventory of multiway analysis techniques and their synergy with compressed deep learning models. The second section provides an in-depth literature review, while the third section introduces a tensorization framework tailored to enhance the expressive power of multidimensional data. Subsequently, the fourth section elucidates the tangible benefits of adopting these approaches through illustrative case studies drawn from tensorized machine learning and deep learning literature. Then, the conclusion section summarises key concepts, delineates existing challenges, and outlines promising avenues for future research in the evolving landscape of tensorization, multiway analysis, and their integration with deep learning models.

## 2. Literature Review

This section reviews the literature on multiway analysis methods, multiway dataset sources, and the tensorisation methods of 2-dimensional datasets. The tensorisation of data given in 2-dimensional format can be as simple as reshaping the data into an n-d array of n being equal to any higher order value to represent each mode independently. For example, a dataset of student grades in all school grades, all cohorts over many years, all subjects, and all exams per subject are usually given in the 2-D format as shown in Table 1.

*Table 1: MATRIX FORM OF HIGH DIMENSIONAL DATASET*

| Academic Year | School Grade | Student ID | Subject | Assessment 1 | Assessment 2 | ... |
|---|---|---|---|---|---|---|
| ⋮ | | | | | | |
| 2021 | 1 | 12345 | Math | 50 | 60 | |
| ⋮ | | | | | | |
| 2022 | 1 | 14734 | Science | ... | ... | |
| ... | | | | | | |

The most intuitive tensorisation of this dataset is a reshape such that mode 1 is the academic year (the cohort), mode 2 is the school grade, mode 3 is the student, and mode 4 is the subject. We can also include each assessment in isolation, but we can take the average or the final assessment as the aggregate for this 4-mode 4-D array. This mapping is from the 2-dimensional coordinate space to the 4-d coordinate space. Coordinates are defined by their basis vector, which defines the unit step into the coordinate. For instance, the academic year is expected to have unit bases of one year per basis. This might be all required for this kind of data from the year the school was established to the current year. However, the student ID might not be incremented by 1 for every new student and might be randomised for each cohort or re-used across the cohorts. For the student ID, hashing into sequential values might be required for a given dataset. Some datasets have minimum and maximum values of acceptable range, and a transformation with respect to the rate of change will be required. For example, for ten academic years, nine school grades, 100 unique students as they level up from a school grade to the next, ten subjects, this will create a Tensor $\mathcal{X} \in \mathbb{R}^{10 \times 9 \times 100 \times 10}$. We can retrieve a specific student record in Python Numpy structure, but identifying the student index i, and retrieve all other modes as $\mathcal{X}[:,:,i,:]$.

Other deterministic and stochastic approaches are discussed in the literature and surveyed (Debals and De Lathauwer, 2015). These will be further explained below.

## 2.1 Multiway Analysis Methods.

The multiway analysis methods take as input a tensor of n-mode and apply various factorisation, regression, clustering, or completion of missing values algorithms. The application of these algorithms is themselves Machine Learning (ML) algorithms that provide insight into the dataset, not just data pre-processing steps. However, they can also be used as a data pre-processing step to better represent the dataset before applying an ML or Deep Learning (DL) model that might be tensorised itself or not. Chapter three of (Helal, 2023) presents the foundation of Multilinear analysis that can be summarised in  Table 2.

*Table 2: Multilinear Algebra & Tensors*

| | Manifolds | Hilbert Space | Curves | Riemannian Geometry | Differential geometry on Manifolds |
|---|---|---|---|---|---|
| Geometric Space | Collection of points and not vectors, on a local Euclidean space forming a Smooth Curved Spaces to Capture Intrinsic Geometry of the data | Hilbert Space $\mathcal{H}$ is a generalisation of Euclidian space in the infinite dimension. | Non-Euclidean spaces using hyperbolic and elliptic geometry to explain Paths or Trajectories | Riemannian Manifolds describe curvatures in higher dimensions and provide geometric properties to facilitate the partial differential equations used in many ML algorithms. | Mapping the data to manifolds with curvature by change of basis using the Jacobian matrix determinant. <br><br> The tangent vector space with basis $\frac{\delta}{\delta x^i}$ for every dimension i, and the dual (cotangent) space with basis $e^i(v)$ / $dx^i$ as the coordinate function (projection on the coordinates) |
| Applications | Dimensionality Reduction, Data Visualization | Maximize Separation, Orthogonality for applications such as Quantum Mechanics, Function Approximation | Capture Curve Shape for applications such as Motion Tracking, Signature Recognition | Preserve Intrinsic Geometry for applications such as Robotics, Shape Analysis, Computer Vision | Capture Local Geometry for applications such as Shape Analysis, Medical Imaging, Computer Vision |
| Similarity or Distance Measure | Various metrics such as: Geodesic Distance, Intrinsic Metric, and even the Euclidean distance measures quantified by Root Mean Square Error (RMSE) | Inner Product using a Kernel function K(xi, xj) = (Φ(xi). Φ(xj)) produces a similarity metric between the data points without explicitly mapping every vector in the dataset. This reduces searching the large space $\mathcal{H}$ to just finding the optimal values of the m coefficients $\alpha_1, \ldots, \alpha_m$ of the features $x_1, \ldots, x_m$. Example Kernel functions include Gaussian radial basis function (RBF) equation and 2-layer sigmoidal neural network. | Arc Length, Frechet Distance such as in solve higher polynomial equations. This can be done in Tensor Form, using tensor metric, which is also a dot product for tensors. | Geodesics, Riemannian metric such as the Fisher metric | Intrinsic Metrics, Curvature Information . This can be achieved by Differential Forms: (Tensors) <br><br> Zero forms (takes a scalar and produce a scalar), 1-form takes a vector and produce a scalar (vector length). 2-forms takes two vectors and produce the area scalar of the parallelogram formed by the two vectors, 3-form takes 3 vectors and produce the volume scalar .... |
| Mathematical Reference First Reference | (Riemann, 1868) | (Hilbert, 1898) | (Fréchet, 1906) | (Riemann, 1868) | (Gauss, 1828) |
| Machine Learning Earlier Adoption | (Tenenbaum, Silva and Langford, 2000) | (Cover and Hart, 1967) | (Gavrila, 1999) | (Fletcher *et al.*, 2004) | (Pennec and Thirion, 1995) |

The first column describes the mathematical structure of the Manifold Learning algorithms that remain in Euclidean space and focuses on finding a lower dimensional representation of a given dataset. The second column is the Kernel Trick that maps the dot product as the similarity measure on two vectors, operating in the infinite dimension Hilbert Space $\mathcal{H}$, using a kernel function that does not do the actual mapping to the higher dimension. This is useful when the dataset is not linearly separable in the lower dimension and can be linearly separable in the higher dimension. The separation linear hyperplane in the higher dimension will be non-linear when projected back to the lower dimension. The third column employs the curved coordinates employing polynomial functions of quadratic degree or higher to fit a dataset. This method increases the number of features to study the interactions by learning the weights of the different polynomial combinations of the given features. This can be as simple as using the Python sklearn.preprocessing.PolynomialFeatures function to apply the regression on the transformed features. Global polynomial optimisation is achieved in Tensor form using tensor decomposition approaches (Marmin, Castella and Pesquet, 2020). This is useful for various function approximation applications for multilinear functions, non-linear functions with polynomials, matrix-matrix multiplication, and systems of polynomial equations. The fourth column expands on the curvatures to map the dataset onto curved coordinates in the higher dimensions introduced by the German mathematician Bernhard Riemann. The Dual space mapping enables applying differential geometry on a coordinate-free approach. The latter uses the Jacobian matrices of the mapping functions to transform the coefficients in one coordinate system/basis to another, and the Hessian matrices, which is the Laplacian of a function to measure the function's curvature as the divergence of its gradient invariant of change of basis. The last three columns use tensors that can benefit from the tensorisation step explained in this paper, and then the multiway analysis methods to provide further insights.

The following is an overview of prior surveys and approaches, focusing on the seminal work by (Kolda and Bader, 2009).

### 2.1.1 Tensor Decompositions

Matrices are factorised using various methods to reduce their dimensionality and identify dominant factors. Projective Methods such as PCA, SVD and others are commonly used, but they lose the non-linear structure of the data. Embedding learning methods keep the non-linear structure while mapping a dataset to a lower dimensional embedding without learning the manifold, such as Multidimensional scaling (MDS). Other methods also learn the manifold, such as Isometric Feature map (Isomap), Locally Linear Embedding and Spectral Clustering. These methods do not work on higher dimensional datasets. Chapter two (Helal, 2023) summarises the algorithmic details of these methods with Python examples. The linear algebra foundations of these algorithms are summarised in Chapter One. The following multidimensional factorisation/decomposition methods are used in high-dimensional datasets, which are explained in chapter 4 of the same book, with the mathematical foundations in chapter three.

### 2.1.1.1 CANDECOMP/PARAFAC (CP) Decomposition

CP decomposition is the multiway extension of SVD and is usually implemented in various constraints and approaches. SVD of dataset X is computed as $X = USV^T = \sigma_1 u_1 v_1^T + \sigma_2 u_2 v_2^T + \cdots + \sigma_r u_r v_r^T$. This can be expressed as the summation of outer products of the vectors of the most dominating columns in U and most dominating rows in V, in the order of the singular values $\sigma$ from the highest $\sigma_1$ to the lowest given rank $\sigma_r$, which are diagonalised in S. This enables the approximate reconstruction of matrix χ as $\hat{\chi}$ from its dominant components: $\hat{\chi} = \sum_{j=1}^{r} \sigma_j u_j v_j^T$. The N dimensions generalisation is defined as $\chi \epsilon \mathbb{R}^{I_1, I_2, \ldots, I_N} \approx [\![\lambda; A^{(1)}, A^{(2)}, \ldots, A^{(N)}]\!] = \sum_{r=1}^{R} \lambda_r \, a_r^{(1)} \circ a_r^{(2)} \circ \ldots \, a_r^{(N)} = \Lambda \times_1 A^{(1)} \times_2 A^{(2)} \ldots \times_N A^{(N)}$., where $\Lambda \epsilon \mathbb{R}^{r, r, \ldots, r}$ is a diagonal core tensor of rank r such that $\lambda_r = \Lambda_{r,r,\ldots r}$. Figure 1 illustrates 2D and 3D SVD.

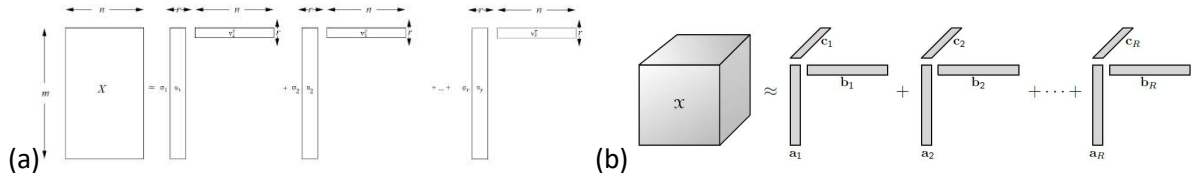

*Figure 1: (a) 2-D SVD, (b) 3-D SVD / CP generalises to the higher dimension (Kolda and Bader, 2009)*

### 2.1.1.2 Tucker Decomposition

The Tucker decomposition is the most cited and is considered a higher-order (or multiway) PCA. It decomposes a tensor into a core tensor (not a diagonal core tensor of weights as in CP decomposition) multiplied by a factor matrix along each mode. PCA for a dataset $x$ can be calculated using a direct projection matrix such that $y = U^T x$, where $y \in \mathbb{R}^p$ is the projected data, $U \in \mathbb{R}^{m \times p}$ is the projection matrix containing the p Eigenvectors and $x \in \mathbb{R}^m = (x_m - \bar{x})$ is the centred m-dimensional dataset standardised to zero mean. PCA can also be calculated using an iterative constraint optimisation method, using a scatter matrix $S_T = XX^T$. The first component is calculated as $\widehat{u_1} = u_1^T S_T u_1 - \lambda(u_1^T u_1 - 1)$, where the Lagrange multiplier $\lambda$ accounts for the normalisation constraints that the principal component should be a unit vector. Then, we minimise by differentiating with respect to $u_1$: and set to 0, $\frac{\delta \widehat{u_1}}{\delta u_1} = S_T u_1 - \lambda u_1 = (S_T - \lambda I) u_1 = 0$, where $\lambda$ and $u_1$ are an Eigenvalue and its corresponding Eigenvector of $S_T$. Iterate through the required components while adding new constraints that each new component is perpendicular/orthogonal to all previous components, and repeat the differentiation step. For example, the second component would be: $\widehat{u_2} = u_2^T S_T u_2 - \lambda(u_2^T u_2 - 1) - \mu(u_2^T u_1)$. The project matrix is the assembled columns of $\widehat{u_i}$ components. The project data is just the simple multiplication $y = U^T x$ (Burges, 2009).

Tucker decomposition for a given tensor dataset $\chi \in \mathbb{R}^{I_1, I_2, \ldots, I_N}$ is computed as $\chi \approx [\![\mathcal{G}; A^{(1)}, A^{(2)}, \ldots, A^{(N)}]\!]$, which performs mode-n multiplication of $\mathcal{G} \in \mathbb{R}^{R_1, R_2, \ldots, R_N}$ as the core tensor such that the values $R_k$ denotes the rank along the $k^{\text{th}}$ mode, and $A^{(k)} \in \mathbb{R}^{I_k, R_k}$ as the orthogonal factor matrices for each mode k. This is illustrated in Figure 2.

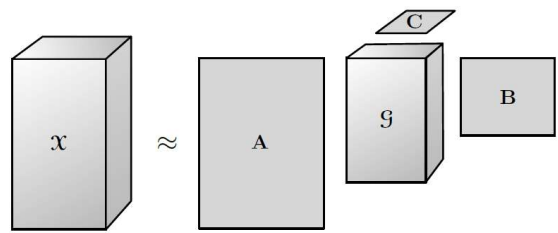

*Figure 2: Tucker Decomposition (Kolda and Bader, 2009)*

### 2.1.1.3 Tensor Networks

Tensor networks is a tensor decomposition algorithm that represents large-scale tensors hierarchically using lower-rank core tensors. The lower-order core tensors are the dominant components in the large-scale tensorial dataset. We can work with networks of tensors, such as each tensor representing a multiway dataset; particular indices/features in a tensor connect to other indices/features in another multiway dataset tensor representation as summation indices, enabling contraction or left as free indices in the final tensor shape. The final tensor shape is the dataset a machine learning or deep learning algorithm should use, identifying some indices/features as predictors and others as target/outcome variables. There are various approaches to producing tensor networks, such as Matrix Product State (MPS) /Tensor Train (TT), Tensor Ring (TR), Matrix Product Operator (MPO), Tree Tensor Network / Hierarchical Tucker, Projected Entangled Pair States (PEPS), and Multi-scale Entanglement Renormalization Ansatz (MERA). For example, the TT decomposition is: $\chi = \mathcal{A}_1 \times_{3,1} \mathcal{A}_2 \dots \times_N \mathcal{A}_N$, where $\mathcal{A}_n \epsilon \mathbb{R}^{R_{n-1}, I_n, R_n}$, $R_0 = R_N = 1$; n = 1, 2, ..., N, such that $\mathcal{A}_1$ and $\mathcal{A}_N$ are of lesser rank than the internal core tensors. Another example is the TR decomposition computed as $\chi = \Re(\mathcal{A}_1, \mathcal{A}_2, \dots \mathcal{A}_N) = \sum_{\alpha_1, \alpha_2, \dots \alpha_N = 1}^{R_1, R_2, \dots R_N} a_1(\alpha_1 \alpha_2)^{\circ} a_2(\alpha_2, \alpha_3)^{\circ} \dots {}^{\circ} a_N(\alpha_N, \alpha_1)$, where the TR rank is defined for each mode as ($R_1$, $R_2$, ..., $R_N$), core tensors $\mathcal{A}_n \epsilon \mathbb{R}^{R_n, I_n, R_{n+1}}$, n = 1, 2, ..., N, and last one $\mathcal{A}_N \epsilon \mathbb{R}^{R_n, I_n, R_1}$ completes the circular connection with the first one, such that $R_1 = R_{N+1}$. These are illustrated in Figure 3.

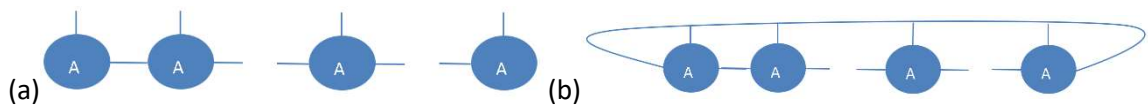

*Figure 3: (a) TT decomposition, TR decomposition*

Other Tensor decomposition methods include Non-negative Tensor Factorization (NTF), INdividual Differences in SCALing (INDSCAL), CANonical Decomposition with LINear Constraints (CANDELINC), PARAFAC2, DEDICOM, PARATUCK2, Hierarchical Tucker(HT), Tree Tensor Network States (TTNS) among others in the literature. All these methods require estimating the rank. Finding the tensor rank is an NP-hard problem, and methods like Alternating Least Squares (ALS) are used. The method is based on fixing A and B to solve C, then fixing A & C to solve B, then fixing B and C to solve A, and repeating until convergence, which is defined as the case when the error is not significantly decreasing. ALS reduces a non-convex optimisation problem to convex subproblems.

## 2.1.2 Tensor Completion

Tensor completion is an extension to the matrix completion class of problems that aims to interpolate missing values in a dataset from the given values. A 3-way association tensor completion application example is presented in (Huang et al., 2021). The dataset used in this paper is collected from the HMDD (the Human microRNA Disease Database), which is a database that curates biological lab experiment-supported evidence for human microRNA (miRNA) and disease associations. Auxiliary data are the disease descriptors collected from Medical Subject Heading (MeSH), a comprehensive controlled vocabulary thesaurus about life science, to calculate disease semantic similarity. Previous work focused on predicting whether a miRNA-disease association exists or not (binary classification/prediction). Instead of building a binary graph association, this paper presented the dataset in a 3-way structure of miRNA-disease-type triplets as a tensor. It introduced Tensor Decomposition methods to solve the prediction task, such that the type explains the roles of miRNAs in disease development or identification. The paper proposed a novel method, Tensor Decomposition with Relational Constraints (TDRC), incorporating biological features (miRNA-miRNA similarity and disease-disease similarity) as relational constraints to further the existing tensor decomposition methods.

They formulated the data as a set of miRNAs $\mathcal{E} = \{e_1, e_2, \dots, e_m\}$, a set of diseases $D = \{d_1, d_2, \dots, d_n\}$ and a set of association types $\mathcal{R} = \{r_1, r_2, \dots, r_t\}$, the authors constructed a multi-relation bipartite graph $\mathcal{G}$. A triple $(e_i, d_j, r_t)$ as a link in the graph $\mathcal{G}$ denoting an association between the miRNA $e_i$ and the disease $d_j$ with the type $r_t$. $\mathcal{G}$ is a binary three-way tensor $\mathcal{X} \in \{0,1\}^{m \times n \times t}$ with miRNA mode, disease mode, and type mode, where each slice is the adjacency matrix with regard to a type of miRNA-disease association. The tensor X is extremely sparse, with many unknown entries, and thus, reaching the goal only by using known links is challenging. The auxiliary data, such as miRNA-miRNA functional similarity matrix and disease-disease semantic similarity matrix, were used to constrain the tensor completion performed by the CP tensor decomposition method. This is an example of integrating multiple datasets that would form a sparse large tensor and constrain the size with known information from other datasets.

### 2.1.3 Tensor Regression

Regression is a supervised ML algorithm that fits a dataset to a mapping function from the features x to the target y by learning the weights of the features that reduce the residual error. The 2-D common linear regression equation is $y = \epsilon + \sum_{i=0}^{N} w_i x_i$, such that the w is the estimated regression parameters, including $w_0$ as the bias and $x_0 = 1$. It can be non-linear using polynomial feature scaling, as mentioned above. It can also use any other non-linear parameters such as exponential, trigonometric, and power functions. Other variants include a multivariate regression model, which is when x is multiple predictors and y can be multiple responses. There are also non-parametric regression models, such as Gaussian Processes (GP), Artificial Neural Networks (ANN), Decision Trees, and Support Vector Regression (SVR). These methods usually require much more sample data than parametric methods (Hou, 2017).

A simple tensor regression model is defined as: given an Nth-order tensor $\chi \in \mathbb{R}^{I_1, I_2, I_3, \dots, I_N}$, and the output $\mathcal{Y}$ could be a tensor of any order required to represent the dependent variable(s); the regression equation is: $\mathcal{Y} = f(\mathcal{X}) + \epsilon$

The f function for linear regression can be the dot product defined in the generalised linear tensor regression model as: $\mathcal{Y} = \langle \mathcal{X}, \mathcal{B} \rangle + \epsilon$

Such that the dot product of the predictor and $\beta$ as the coefficient tensor in the same size as the predictor $\chi$, capturing its tensor covariate, and is added to $\epsilon$ as the tensor representation error or bias. Similar non-linear functions in the higher order can be employed. Prediction or reconstruction/interpolation can occur based on a dataset of M samples as follows: $\hat{x}_{i_1, i_2, \dots, i_N} = \sum_{k=1}^{M} \langle \chi_k, \beta_k \rangle$.

Solving $\langle \mathcal{X}, \mathcal{B} \rangle$ by vectorising, both tensors will produce a huge number of parameters. For example, an MRI dataset $\mathcal{X} \in \mathbb{R}^{128 \times 128 \times 128}$ will require 2,097,152 + and five usual covariates parameters to estimate, which is intractable. Using the unsupervised PCA produces the most dominating principal components that are irrelevant to the input, lose the multiway structural relationship, and are difficult to interpret. The CP Tensor Regression defines the $\mathcal{B}$ tensor in terms of its rank-R CP decomposition, a $[B_1, B_2, \dots, B_N]$ with $B_n = \left[ b_1^{(n)}, \dots, b_R^{(n)} \right] \in \mathbb{R}^{I_n, R}$, such that $y = \langle \mathcal{X}, \mathcal{B} \rangle + \epsilon = \left\langle \mathcal{X}, \sum_{r=1}^{R} b_r^{(1)} \circ b_r^{(2)} \circ \dots b_r^{(N)} \right\rangle + \epsilon$ where y is a scalar output. This reduces the number of parameters from $O(I^N)$ to the scale of $O(NIR)$ while also producing reasonable reconstruction accuracy. For example, the previous MRI example parameters can be reduced to $389 = 5 + 128 \times 3$ for a rank-1 model and to $1,157 = 5 + 3 \times 128 \times 3$ for a rank-3 model.

Tucker decomposition is more flexible than CP and accurately captures the multiway structural relationships in the core tensors. Tucker Tensor Regression defines the $\mathcal{B}$ tensor in terms of its Tucker decomposition as:

$$\sum_{r_1=1}^{R_1} ... \sum_{r_n=1}^{R_N} ... g_{r_1...r_n} b_{r_1}^{(1)} \circ b_{r_2}^{(2)} \circ ... b_{r_n}^{(N)}$$

such that $y = \langle \mathcal{X}, \mathcal{B} \rangle + \epsilon = \left\langle \mathcal{X}, \sum_{r_1=1}^{R_1} ... \sum_{r_n=1}^{R_N} ... g_{r_1...r_n} b_{r_1}^{(1)} \circ b_{r_2}^{(2)} \circ ... b_{r_n}^{(N)} \right\rangle + \epsilon$ where $\mathcal{G} \in \mathbb{R}^{R_1, R_2, ..., R_N}$ with entries $\{g_{r_1...r_n}\}_{r_1=1,...r_n=1}^{R_1,...,R_N}$. The factor matrices are defined as $B_n \in \mathbb{R}^{I_n, R_n}$ along different modes. This reduces the number of parameters from $O(I^N)$ to the scale of $O(NIr + r^N)$, which is higher than the CP regression parameters of $O(NIR)$ but more parsimonious modelling of the input data when R $\ll$ $N$. An example application presented in (Li, Zhou and Li, 2013) shows that for a tensorial dataset representing neuroimaging data as a 3D signal $\mathcal{X} \in \mathbb{R}^{16 \times 16 \times 16}$ using a Tucker model with multilinear rank = (2, 2, 5), the number of parameters is 131, while using a 5-component CP regression model yields 230 parameters.

### 2.1.4 Tensor Clustering

Clustering is an unsupervised machine learning approach such that given an unlabelled data matrix X, it can be represented as X≈AB$^T$, such that each row in A (the canonical basis vector) selects a row in B, which contains the clustering vectors. Estimating A and B (two unknowns) from X enables multiway clustering. Dictionary Learning Algorithms and Source Separation Algorithms, such as Independent Component Analysis, attempt to estimate two matrices for the given data matrix, among others, are examples of unsupervised ML algorithms. All these algorithms are expanded to multiway higher dimension formulation (Acar and Yener, 2009).

## 2.2 Multiway (Tensorised) dataset sources

This section surveys the datasets to use for multiway analysis. The dataset can be tensorised from vector or matrix forms. Tensors can be formed as well by integrating multiple datasets using various data fusion algorithms. However, many data sources are available in tensor form as well.

### 2.2.1 Traditional Datasets

Kaggle and UCI, among other public domains, are repositories containing massive amounts of datasets on various application domains and ready to use for various ML algorithms. The vast majority of these datasets are 2-dimensional in nature and are ready for the applications of linear algebra-based ML algorithms. Data fusion techniques can combine the relevant datasets into a multi-modal ML model to create a high dimensional tensor of the various modes to be the input of a multiway multi-modal ML algorithm (Baltrušaitis, Ahuja and Morency, 2017). The process of tensorisation requires understanding the different modes in one 2-D dataset or collected from multiple 2-D datasets. This analysis of the modes helps to decide how to integrate the modes from different datasets or divide a mode from one dataset into different modes as the application requires. Dividing a mode into more modes is called segmentation (Debals and De Lathauwer, 2015). An example of integration is hospitals' trial statistics, which might have a common mode between the different hospital datasets to integrate on. An example of dividing a column from one dataset into two modes is division by year from a date column and having a different mode for each year, gender, age groups or others.

### 2.2.2 Graphs or Networks Datasets

Sensor networks collect high-dimensional datasets naturally. For example, (Almomani, Al-Kasasbeh and AL-Akhras, 2016) proposed a simulated dataset for Wireless Sensor Networks (WSN) that contains sensors in different locations that gather various information and send them in messages collaboratively to Base stations. These sensors usually have limited energy and memory and need to optimise the communication of messages using clustering algorithms such as the Low Energy Aware Cluster Hierarchy (LEACH), which is a routing protocol that optimises the energy consumption by organising sensor nodes into clusters to distribute the energy among all nodes in the network. This dataset can form a Tensor $\mathcal{X}^{N,B,K,M}$ for N sensors, B Base Stations, K Clusters of sensors, and M Messages between sensors and other sensors, Base stations, or cluster heads.

Knowledge Graphs are also naturally high-dimensional datasets. These graphs usually employ RDF triplets of two entities and a relationship. The entities can be of various types, creating a mode per type, and similarly, the relationship can be of various types, and create further modes for the different types. A bibliography network can contain authors as the main entity type and co-authoring as the relationship. This would create a Tensor $\mathcal{X}^{n \times n \times m}$, of n entities x n entities x m relationships. Various datasets are considered knowledge graphs, such as:

- The Citation network in the Cora dataset introduced by (McCallum et al., 2000),
- Kinships in Australian Tribes (Dousset et al., 2010),
- Nations clusters of countries, clusters of interactions between countries, and clusters of country features such as indicators selected from the (The World Bank, 2023).
- UMLS (Unified Medical Language System) (Bodenreider, 2004).
- Semantic Web's Linked Open Data (LOD) contains millions of entities, hundreds of relations and billions of known facts (Khusro, 2014).
- YAGO 2 ontology contains 4.3x1014 possible triplets as a knowledge base about people, cities, countries, movies, and organisations (Hoffart et al., 2013).
- Dbpedia ontology is an open knowledge graph crowd-sourced from the web contents to create structured knowledge such as the Dbpedia-Person dataset (Mendes et al., 2011).
- Multi-modal datasets include the Visual Question Answering (VQA) dataset (Goyal et al., 2017) and Multi-modal sentiment analysis (Das and Singh, 2023), among others.

### 2.2.3 Image and Video Datasets

The famous example of the MNIST dataset of handwritten digits contains images of $28 \times 28$ pixels, usually flattened as 784 pixels in one-row values per image, representing grey shade as integers between 0 and 255. If kept in a 2D form, the covariance matrix will be smaller. This is the reason why 2D Convolutional Neural Networks (CNN) are much faster and more accurate than Dense Layers for spatial datasets like images. Data in the tensor form can represent a coloured image with pixel values for red, green, and blue as three different values in the (RGB) frames stacked into a 3rd-order tensor. Similarly, a video dataset can include the 3rd-order coloured image frames extended with the time dimension in a 4th-order tensor. For a video example illustrated in Figure 4, the first two dimensions are spatial rows and columns of 128 x 88 dimensionality and a time third dimension of 20 frames. A Linear Subspace Learning (LSL) vectorisation in (a) performed by the product of the number of dimensions in each mode results in a large covariance matrix of 189 GB memory fingerprint and the resulting processing time. On the other hand, a Multi-linear Subspace Learning (MSL) tensor-based analysis performing the sum of three smaller covariance matrices results in 95.8KB of memory fingerprint and reduced processing time.

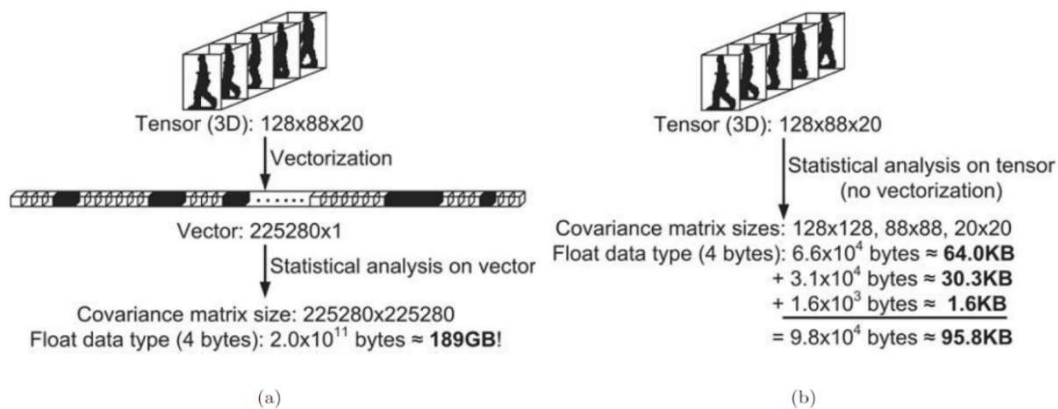

*Figure 4: LSL vs MSL using tensorised data compression (Lu, Plataniotis and Venetsanopoulos, 2011)*

Possible video datasets that can create higher tensors by adding other modes for other ML tasks include the Human limb motion dataset. These datasets collect videos about persons doing specific movements, using video cameras detecting the 3D position of infrared markers placed on each person's legs, arms or anywhere. A similar dataset was collected by (Della Santina *et al.*, 2017). Tensors can be created for the regular 4-way video recordings, adding modes for a specific person, specific sensor, specific motion, or any other scene analysis objectives. Another example is (Soliman *et al.*, 2019) dataset that contains 1000 violence and 1000 non-violence videos collected from YouTube videos. The violent videos have many real street fight situations in several environments and conditions. The non-violence videos are gathered from various human actions like sports, eating, walking, and so forth.

### 2.2.4 Health and Biomedical Datasets

Brain-computer interface (BCI) based on EEG signals are naturally multi-mode due to the data recording mechanism. For example, signals are recorded by multiple sensors (electrodes) in multiple trials and epochs for multiple subjects and with different tasks, conditions…, and so forth. This dataset can be represented in rank n tensors to enable multi-way, multi-block data analysis techniques.

Magnetic resonance imaging (MRI), functional MRI, PET, and MEG datasets are also naturally multi-mode. For example, A NIfTI file for a typical MRI scan stores the voxel values in an array of numbers. The coordinates for a single voxel within a NIfTI image volume can be specified as a 3-dimensional index (x, y, z) or a 4-dimensional index (x, y, z, t) for time. Then the subject is another mode, then the aim of the experiment is another, the resolution and so forth. Similar datasets can be found at OpenNeuro (Markiewicz *et al.*, 2021).

### 2.3 Tensorisation Methods.

Tensorization is the process of creating multidimensional datasets from existing 2-dimensional datasets or generally transforming a lower dimensional array into a higher dimension array by merging multiple datasets and segmenting one dataset. This can be simple reshaping and can be using various deterministic and stochastic methods. This step needs to be carefully considered if data collection is done from scratch.

### 2.3.1 Reshaping

Given a dataset in matrix form, tensorisation can be simple multiple indexing, pivot table transformation, or actual transformation into the higher space. The accompanying source code shows examples of public domain datasets tensorised using different methods. The first data set is Global Temperature for 100 cities records from 01/11/1743 to 01/09/2013 (Rohde and Hausfather, 2020). There are 239177 rows, such that the number of records per city is different. Only the date and city can be coordinates as they contain different information. Country, latitude, and longitude columns are unique for each city, and converting them to new coordinates is not reasonable. The second dataset (*Gender Pay Gap Dataset*, no date) tensorizes wages per gender, region, age, degree, and occupation up to rank-5 tensors. Table 3 summarises the different data structures and estimates the memory size and sparsity.

*Table 3: Different Data structures compared with tensorized data*

| Data | Pandas Indexing | Pivot Table | Tensor |
|------|-----------------|-------------|--------|
| Order-one tensor, temperature per city | 239177 rows × 2 columns = 478,354 | 100 rows for 1 index column | (100) − 100% dense, no need to sparsify, quantise over cities is not useful |
| Order-two tensor, temperature per city and latitude | 239177 rows × 3 columns = 717,531 | 100 rows for 2 indices columns | (100, 99) = 9,900 Only 2 values are zero, this is dense, and sparsifying is not useful, and quantisation of cities and dependent latitude is not useful. |
| Order-three tensor, temperature per city, latitude and longitude | 239177 rows × 4 columns = 956,708 | 100 rows for 3 indices columns | (100, 99, 272) = 2,692,800 1% dense and 99% sparse. Using sparse arrays saved 94% of the memory of nd-array and > 99% of pandas frames memory. |
| Order-four tensor, temperature per city, latitude, longitude and date − all the dataset | 239177 rows × 5 columns = 1,195,885 | 228175 rows for 4 indices columns | (100, 99, 272, 3239) = 8,721,979,200 values Sparse as only 239177 non-zero values are found in the dataset. We can also quantise for the 271 years only by averaging the temperature for each year for every city, or for each month for seasonality |
| Order-one tensor, wage per gender | 33398 rows × 2 columns | 2 rows for 1 index column | 2 values |
| Order-two tensor, wage per gender and region | 33398 rows × 3 columns | 10 rows for 2 indices columns | (2, 5) = 14 values |
| Order-three tensor, wage per gender, region and age | 33398 rows × 4 columns | 323 rows for 3 indices columns | (2, 5, 40) = 400 values with 323 non zero |
| Order-four tensor, wage per gender, region, age, and degree | 33398 rows × 5 columns | 957 rows × for 4 indices columns | (2, 5, 40, 3) = 1,200 values with 957 non-zero |
| Order-five tensor, wage per gender, region, age, degree, and occupation | 33398 rows × 6 columns | 21829 rows × for 5 indices columns | (2, 5, 40, 3, 975) = 1170000 values with 21845 non-zero. This is 99.4% sparse to use only 4.44% memory |

**Sparse Structures:** This case-by-case tensorization showed the critical need for sparse arrays. scipy.sparse Python package offers data structures that work for 2-dimensional arrays only. The source code showed possible attempts to create sparse arrays using scipy.sparse data structures if they advance their implementation to nd-arrays.

Meanwhile, a dictionary data structure of n-dimensional indices tuple as key and aggregated value pairs are created for sparse n-dimensional arrays. The problem of custom data structures will also require rebuilding custom linear algebra operations and ML algorithms. Current 2-dimensional ML algorithms advise against sparse datasets and use dimensionality reduction to remove the sparsity. However, it can be very useful to show the multiway interactions between the different features.

**Quantisation:** It is also important to quantise while doing the coordinate change as a safe model reduction method. Choosing the most suitable basis will affect the performance of the ML algorithm in memory, speed and accuracy metrics.

**Parallelisation:** The work of (Helal *et al.*, 2008, 2009; Helal, 2009) implemented Tensor Partitioning on a cluster of computing nodes as an example of wavefront processing of N-D arrays applied to the Multiple Sequence Alignment problem. The ND arrays were expressed dynamically using a C data structure containing Ndim, shape, and data as parameters and creating a linear array in memory. The N-dimensional index is then parameterised to access a specific location in the array or update it. The dense partitions are only created in the memory of the computing processor from a specific index up to a partition size to be processed independently on different processors. For example, shape (100, 100, 100, 100, 100) and partition size (10, 10, 10, 10, 10) will have the first wave computing on one processor the first partition from (0, 0, 0, 0, 0) to (9, 9, 9, 9, 9). Wave 2 will divide between the available processors the available partitions with starting indices that sum up to 10, such as (10, 0,0,0,0), (8, 1,1,0,0), (7, 1,1,1,0), … etc., and all permutations and number combinations. The processor ID identifies its share of the nd-indices in the current wave and the dependent partitions from previous waves of computation to receive its scores for direct communication. This model was done in C using MPI that can run over multiple cores or clusters of computing nodes over any network. This can easily be done in Python using GPUs or other parallel processing platforms.

Numpy, MatLab, and Mathematica enable building arrays of any dimensionality and shape. It is very important not to fix the array dimensionality and shape in the input of ML stacks such as SKlearn, TensorFlow and others. This limits the ability to dynamically create these arrays from the different datasets, particularly in the tensorisation step, as shown in the tensorisation by adding coordinates Python notebook.

### 2.3.2 Deterministic Tensorization

Hankelization and Löwnerization are all deterministic tensorisation methods that systematically project any dataset to a higher dimension. Deterministic methods can be detensorised to 2-D or 1-D, or any lower dimension by applying the reverse process (Debals and De Lathauwer, 2015). Hankelisation creates a skew-diagonal higher order of the original data elements. The data is assumed to be exponential, sinusoidal, or polynomial functions that can be mapped to a tensor shape of a precise rank. This is useful in harmonic retrieval, direction-of-arrival estimation, and sinusoidal carriers in telecommunication for applications such as Blink Source Separation (BSS). This is useful in estimating the subspace in which a high-dimensional dataset may reside. The higher-order parameterised implementation of the Hankelisation procedure is found in (Vervliet *et al.*, 2016) as Matlab functions in the TensorLab package. An attempt to rebuild it in Python is in the accompanying source code, with enough details to assess the BSS example. Comparing the Hankelised higher-order tensor of the mixed signal that is decomposed in the CP algorithm with PCA and ICA performance, the residual error is higher in the multi-way solution, then ICA, then PCA as illustrated in Figure 5. However, the multi-way reconstructed signals are closer in shape to the original signals than both ICA and PCA reconstructed signals as illustrated in Figure 6.

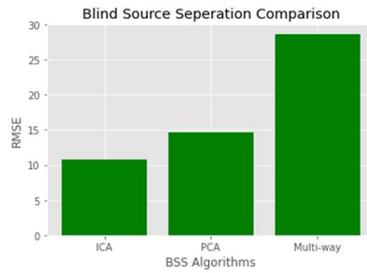

*Figure 5: BSS residual error from ICA, PCA and multi-way decomposition of the hankelised tensor*

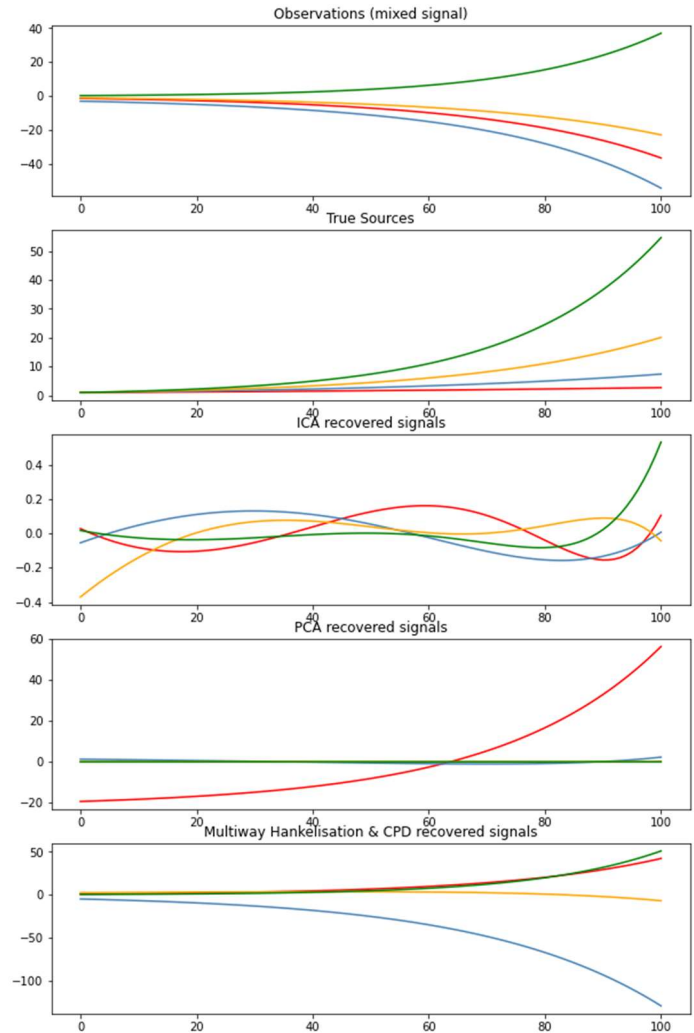

*Figure 6: BSS reconstructed signals from ICA, PCA, and multiway compared to the mixed signal and the original sources.*

Löwnerization is another deterministic approach that maps a given dataset to a higher dimension, assuming the data are of Rational function basis. This is also implemented in (Vervliet *et al.*, 2016) as Matlab functions and its reverse process. Other tensorisation functions implemented in TensorLab include segmentation and decimation and their reverse processes. They are parameterised differently to enable various applications to find a suitable subspace that represents the dataset being analysed.

### 2.3.3 Statistical Tensorization

Second-order statistics, such as the covariance matrix, can be used to tensorise a dataset along particular modes. The covariance matrix is usually a matrix of the variance of each feature (variable or column) with each other, as estimated from N rows or observations. In the higher order, this will be repeated for the higher dimension. This is implemented as well in (Vervliet *et al.*, 2016) as dcov Matlab function generalising the cov function, which identifies the modes of the observation and features and generates the higher order covariance of the features with the remaining modes. Also, lagged second-order statistics are implemented in the same TensorLab package using scov that returns shifted covariance matrices stacked along the third mode. This is parameterised in the lags argument.

For non-gaussian datasets with statistically independent latent variables, the higher-order statistics such as cumulants and moments describe the data distribution. The p-norm of a vector is a positive-definite scalar function defined as $\|v\|_p = \left(\sum_{i=1}^{N}|v_i|^p\right)^{\frac{1}{p}} \geq 0, \forall p \geq 1$, where $|v_i|$ is the absolute value of each element $v_i$. This means that 1-norm is the sum of the absolute values of the elements. The 2-norm is the magnitude of the vector $v \in \mathbb{R}^N$, which is its length (Frobenius norm) and is denoted $\|v\|_2$ or $\|v\|_F$. It is the Euclidean distance from the origin to the point reached by the vector and calculated as follows $= \sqrt{\sum_{i=1}^{N} v_i^2}$. The infinity-norm is defined as the case where $p \to \infty$, as $\|v\|_\infty = \lim_{p \to \infty}\left(\sum_{i=1}^{N}|v_i|^p\right)^{\frac{1}{p}} = \max(|v_i|)$. For example, given $v = \begin{bmatrix} 10 \\ 2 \\ -6 \end{bmatrix}$, then $\|v\|_1 = 18$, $\|v\|_2 = 11.83$, $\|v\|_\infty = 10$. The central moments describe a distribution by its mean of a sample or the Expected value of the population as the weighted average $E(X) = \sum_{i=1}^{N} p_i x_i$ where p is the probability/frequency/weight of the value $x_i$ as the first central moment, the second central moment is the variance, the third is the skewness, and the fourth is the kurtosis. The $n^{\text{th}}$ central moment is $\mu_n := E[(X - E[X])^n] = \int_{-\infty}^{\infty}(x - \mu)^n f(x)dx$. The cumulant of a random variable is calculated in the form of a cumulant generating function, which is the logarithm of the moment generating function as $K(t) = \log E[e^{tX}] = \sum_{n=1}^{\infty} k_n \frac{t^n}{n!} = k_1 \frac{t^1}{1!} + k_2 \frac{t^2}{2!} + \cdots = \mu_t + \sigma^2 \frac{t^2}{2} + \cdots$. The first, second and third cumulants are identical to the first, second and third central moments. They differ at the beginning of the fourth cumulant. Cumulants are easier to compute for their excellent mathematical properties, such as going to zero when variables are independent, and they describe the connectivity of the data, while the central moments do not. Cumulants are also implemented in the same TensorLab package as cum3, cum4, xcum4 and stcum4 Matlab functions. These are also useful in describing non-Gaussian datasets and are often used in BSS using a quadrilinear mapping of the fourth cumulant.

### 2.3.4 Domain-transform Tensorisation and other methods

Domain-transform methods can be used in Tensorisation to represent some signal datasets better. The values stored in a given index vector capture interaction between the dimensions/modes. This means symmetric or partially symmetric tensors might be sufficient to capture the inter-mode interactions, ignoring the values in a permuted index (same modes in a different order) in which the value might be redundant. For example, in the EEG dataset, we have F frequency measures collected over T time samples from S channels, forming a 3rd-order tensor. Transforming the domain of this 3rd-order tensor to get the time-frequency decomposition can be achieved using a short-time Fourier transform (STFT) that uses a fixed window size or wavelet transform (WT) that uses variable window sizes inversely proportional to the frequency resolution (high or low). Other transformations can represent data at multi-scale and orientation levels, such as the Gabor, contourlet, or pyramid steerable transformations. More details about the change of basis and representation learning are discussed in chapter five of (Helal, 2023).

Furthermore, another higher-order statistic to describe a dataset is the partial derivatives of the observations' Generalised Characteristic Functions (GCF). Using a suitable tensor format, such as the lower-rank core tensors presented in section 2.1.1, enables all the above-discussed transformations and others while keeping the number of parameters smaller. The monograph by (Cichocki *et al.*, 2016, 2017) provides detailed discussions with examples of various forms of tensorisation that prepare a dataset for compressed tensorised deep neural network models. Tensorising datasets, for example, by using tensor network representations, often allows for super-compression of datasets as large as $10^{50}$ entries down to the affordable levels of $10^7$ or even less.

## 3. Proposed Tensorisation Framework

The previous section explained various methods that can be used to tensorise a dataset or a collection of datasets in one tensor form. The tensorisation is the first step in applying tensor computing or multi-way analysis. The complete framework is illustrated in Figure 7. The Tensorisation step can be itself a representation learning step. By studying the dataset, Tensorisation might take into consideration the most useful basis functions for the required number of coordinates. However, the interactions of the variables might not be clear enough, and another step of representation learning might be done manually or by dimensionality reduction algorithms such as PCA and SVD for 2-way analysis or Tucker and CPD as the equivalent multi-way analysis. Also, mapping the dataset to the most useful representation might be all the analysis required to perform various ML tasks, such as tensor completion, clustering and classification. It might be a data pre-processing step for another ML algorithm, such as various Deep Learning models. Various Tensorisation occurs at the various steps in building an artificial neural network (ANN) model.

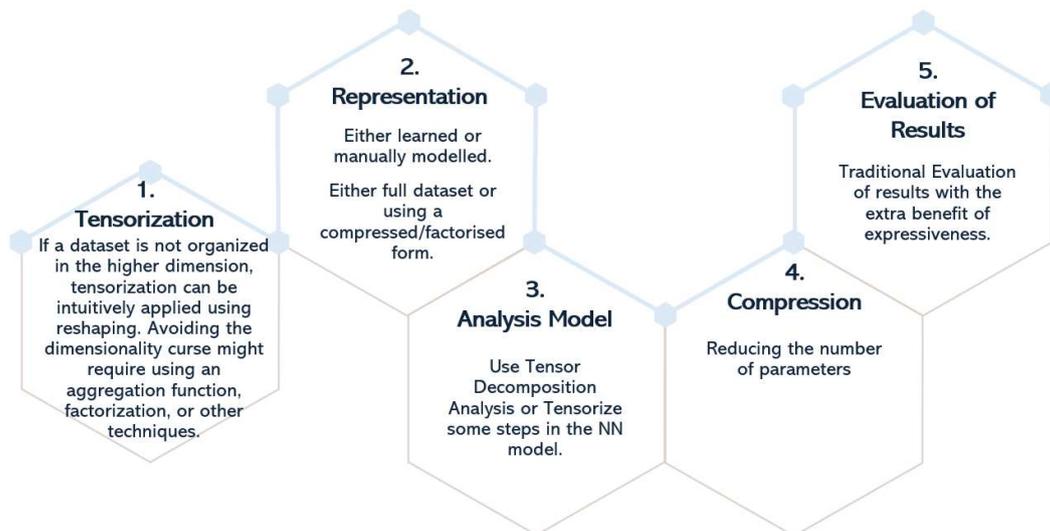

*Figure 7: Tensor Computing Framework*

## 4. Application of Tensorization in Deep Learning Models

In the first ANN building block, activation and loss function choice can be tensorised, as shown in Figure 8. The standard weighted sum of dot products of input vectors with weight vectors and then aggregating them is suitable for vector and matrix form datasets. A tensorised activation function for a tree data structure can start from the node assigned to a neuron to recursively compute the weighted sum of its children with weight sharing between neurons to reduce the model complexity. This model, called a recursive neuron, is first modelled for binary trees and leads to a higher-order generalised n-ary tree using tensorised aggregation. CP decomposition or Tensor Train (TT) decomposition can further decompose the full format tensor aggregation. Chapter six will introduce more details about tensorised activation functions (Bacciu and Mandic, 2020). The loss function can use the decomposed tensor cores of the weights tensor.

In the second ANN building block, the choice of the number of hidden layers and the number of neurons per layer can also benefit from the compressive nature of tensor decomposition algorithms. Tensor Networks can compress the whole Deep Neural Network (DNN) using a suitable tensor decomposition algorithm and then map back to the uncompressed form. Another approach is to update only the final fully connected layer (or specific layers of interest) of a model with a tensor decomposition layer, such as TT. Usually, compressed models benefit from wider, fewer layers (shallower networks).

Low-rank matrix factorization of weights in fully connected layer

$$\underset{(M \times N)}{\boldsymbol{W}} \approx \underset{(M \times r)}{\boldsymbol{A}} \underset{(r \times N)}{\boldsymbol{B}} \qquad \text{Compression}: \mathcal{O}(MN) \to \mathcal{O}(r(M+N))$$

Low-rank tensor network factorization of weights

- ▶ Step 1: $W \to \mathcal{W}$ (Matrix to $d$-order tensor)
- ▶ Step 2: $\mathcal{W} \approx \mathrm{TT}(\mathcal{G}_1 \cdots \mathcal{G}_d)$ (Tensor network representation)

Loss function:

$$\mathcal{L}(\boxed{\boldsymbol{W}}, \boldsymbol{x}, \boldsymbol{y}) \to \mathcal{L}(\boxed{\mathcal{G}_1, \cdots, \mathcal{G}_d}, \boldsymbol{x}, \boldsymbol{y})$$

Compression:

$$\mathcal{O}(MN) \to \mathcal{O}(dr^2 \sqrt[d]{M} \sqrt[d]{N})$$

Order of tensor    TT-rank

Train very "wide" model

*Figure 8: Tensorising Neural Networks (Novikov et al., 2015)*

This new arrangement of the tensorised data will require an alignment of the data slicing into the different epochs. These blocks of tensorised data need to represent the multi-way structures in the dataset identifying latent variables so that the learning iterations can reduce the error.

## 4.2 Tensorising ML and DNN Case Studies and Experiments

The following reviews various experiments with Tensor computing in different domains. More details about these experiments detailing their approaches can be found in (Helal, 2023).

**Signal Processing:**

An example of the BSS problem using tensor decomposition is presented in (Böttcher *et al.*, 2018). The authors built a Python Package, "Decompose", that generalises the PCA, ICA, and NMF solutions to the BSS problem. These methods are built on statistical assumptions that might be found in a dataset or might not be found. Each of them would produce different sources when applied to the same dataset. Expert knowledge is usually needed to identify the correct statistical assumptions of a given dataset in an application domain. The authors built a probabilistic BSS model that estimates the priors of every source, can extend to new prior distributions, scales well to large datasets, assuming each source has a different sparsity level, and efficiently estimates the posterior adapted to the dataset.

**Data Warehousing and Business Intelligence:**

Data cubes are multidimensional data analysis methods provided by various Data warehousing providers. It is based on dividing any dataset into Dimensions and Measures. The dimensions are the coordinates/order of the tensor, and the measures are the values it stores. Any dataset in matrix form can be analysed by transforming it into Data cubes, which are hypercubes in the higher dimension. Various cube algebra operations can be applied, such as slice and dice for partitioning, drill up and down for hierarchical dimensions, aggregation for compression, and pivoting for navigating the different views. Online Analytical Processing (OLAP) aim to pre-compute all required operations for faster response time. This remained computationally prohibitive as the dimensionality increased. The literature has a plethora of approaches to reduce the complexity by compression, approximation, and parallelisation. The work (Peiris, 2017) hypothesizes from experimental results that the CUBE operator of Spark's DataFrame API performs better in distributed cube materialisation than MapReduce.

The work (Spelta, 2017) applied CP tensor decomposition of a 3D-tensor $D \in \mathbb{R}^{N \times N \times Z}$ in which the first and second dimensions are for a basis of one stock out of N stocks, and the third dimension is the time series of length Z. The tensor values $\delta_{klz}$ represents the distance between stock k and stock l at time z. The CP decomposition produces the outer product of three vectors $D \cong \beta v \circ v \circ u$, such that $v \in \mathbb{R}^N$ is the two vectors, including the total spatial dissimilarity between stocks and the vector $u \in \mathbb{R}^Z$ shows the temporal profile of the dissimilarities, and $\beta \cong \|v\| \|v\| \|u\|$. The predicted vectors compose the predicted tensor D as the distance matrix generalisation that connects each stock asset's intensity with the other assets over time. This similarity is computed by summing the distances between each pair of assets, producing a centrality score $\hat{F}_k = \frac{1}{N} \sum l \hat{d}_{kl}$ . A predicted asset stock price change is computed as $\Delta_k^{t+1} = \hat{F}_k^t - \hat{F}_k^{t-1}$. They applied this method to the S&P500 dataset of 388 stock assets with time series prices over 3827 work days and another two datasets: FTSE MIB of 59 and Euronext Paris of 156 stocks, both with prices over almost three years of time series. Their results revealed the increasing trend and three recessions of the dot-com bubble in 2002-03, the financial crises of 2008-09, and a recession around the Greek parliamentary elections. Prior methods used correlation networks that aggregate the data to pair-wise distance matrices, causing loss of crucial information, while the tensor decomposition method relied on the distance evolution over time.

**Computational Science:**

A 3-way association tensor completion application example is presented in (Huang *et al.*, 2021). The dataset used in this paper is collected from the HMDD (the Human microRNA Disease Database), which is a database that curates biological lab experiment-supported evidence for human microRNA (miRNA) and disease associations. Auxiliary data are the disease descriptors collected from Medical Subject Heading (MeSH), a comprehensive controlled vocabulary thesaurus about life science, to calculate disease semantic similarity. Previous work focused on predicting whether a miRNA-disease association exists or not (binary classification/prediction). Instead of building a binary graph association, this paper presented the dataset in a 3-way structure of miRNA-disease-type triplets as a tensor. It introduced Tensor Decomposition methods to solve the prediction task, such that the type explains the roles of miRNAs in disease development or identification. The authors formulated the multi-type miRNA-disease association prediction as a tensor completion task. Their goal was to complete the tensor for exploring the unobserved triple associations using Tensor Decomposition Methods. The authors concluded that TDRC could produce better performance while being more efficient. The reconstructed tensor's low-rank property may help further improve the performance.

This application illustrates a method that can be generalised to tensorize any relational dataset such that several types of relationships are involved.

**Machine Learning:**

The first application of tensors in data mining to review is contributed by Acar et al. (Acar et al. 2005), (Acar et al., 2006), who applied different tensor decompositions to the problem of discussion disentanglement in online public chatrooms on the Internet Relay Chat (IRC), and how social networks evolve. The dataset contains text messages with timestamps, nicknames of identities of sender and receiver, and timestamps of nicknames quit/leave or kick. The nicknames could belong to the same person, pretending to be of any age and gender. The topics' keywords were semantically analysed, and data distributions were estimated. The dataset is multidimensional and noisy. The authors constructed a tensor T of order three, capturing users, keywords, and time in each mode, respectively, such that where $T_{ijk}$ is user i, sent several of keyword j during time slot k. Then Tucker1 and Tucker3 were applied to identify user groups, which are set of users sharing a maximal keyword set in a given time period. This is achieved by an indeterministic c-means clustering algorithm running 100 times, which returns multiple memberships for each data point. They compared to 2-way SVD clustering of user/keyword, and user/timestamps to show that the 3-way was more accurate. Then, they defined a 4-way Tensor to add the IRC server as the fourth mode to identify the computational efficiencies of tensor higher-order representations and decomposition methods compared to the corresponding pair-wise approaches.

The work in (Bader et al., 2008) used CP for automatic conversation detection in the Enron Email dataset over time using an m term x n author x q month 3-way tensor $X^{m,n,q}$. Based on previous literature on the analysis of this dataset, the authors selected an interesting subset of this dataset that makes a tensor $X^{69157, \ 197, \ 12}$ with 1,042,202 non-zeros entries scaled to their weighted frequency. They applied Parafac decomposition using ALS such that $\|X - \sum_{l-1}^{r} \ A_l \circ B_l \circ C_l\|^2$. The decomposed tensor factors were $A^{m \times r}$ containing the highest scores for terms/topics, $B^{n \times r}$ containing the highest scores for authors, and $C^{q \times r}$ containing the highest scores for these topics over time; they chose a 12-month duration. The rank r was chosen to retrieve a specific number of topics in the data, setting it to 25. Eight topics out of the 25 were interpretable in the context of other events happening in the same time duration, as compared to the two-way (term-author) NMF method that could not extract these discussions. The decomposition predicted discussion threads and produced charts of previous focused discussions over time.

The authors of (Nickel et al., 2011) proposed a relational learning approach RESCAL based on the DEDICOM tensor decomposition method with relaxed constraints. They exploited a three-way tensors $X^{n, n, m}$, n entities x n entities x m relationships such that when $X_{ijk} = 1$, this means that entity i has a relationship of type k with entity j. Entities. Domain data is given in the form of Resource Description Framework (RDF) triplets. A tensor decomposition of the form $X$ as $X_k = AR_kA^T$, where A is a n×r matrix containing the latent-component representation of the repeated entities in two modes from the domain and $R_k$ is an asymmetric r×r matrix that models the interactions of the entities in the k-th predicate. The computed low-rank representation of the domain data is used in the prediction of a link as $\hat{X}_{ijk} > \theta$, for some threshold θ. The collective classification can be performed by slicing the low-rank reconstructed tensor for a given class relationship or actually reconstructing the relevant slice only. Also, Link-based clustering of entities can be performed using a similarity measure between entities based on their similarity across multiple relations. The authors compared the performance of RESCAL compared to standard tensor factorisations such as CP and DEDICOM, and relational learning algorithms such as statistical unit node-set (SUNS) and the aggregated SUNs+AG. The conducted various experiments on various datasets. They showed that the results of RESCAL and DEDICOM outperform both CP and SUNS on all datasets.

Another application is the proposal of new ontologies' terms from the data to knowledge database engineers as decision support systems to evolve an ontology by grouping/clustering instances. Out of 87 predicates relation in YAGO 2 at the time, 38 were used to form a sparse tensor $X^{3000417,3000417,38}$ with 41 million entries and a sparse attribute matrix $D^{3000417,1138407}$ with 35.4 million entries. Of the $4.3 \times 10^{14}$ possible triplets in YAGO 2, only $4 \times 10^7$ non-zero entries were available. The authors compared the RESCAL to other tensor factorisation and classical relational learning methods, showing that RESCAL is more efficient in predicting RDF triplets and other machine learning tasks (Nickel et al., 2012).

The authors of (Padia et al., 2016) further extended the RESCAL tensor decomposition method to RESCAL+ approach by adding a similarity matrix to the minimisation algorithm to force slices of the relational tensor to decrease their differences between one another to achieve unequal contribution of the slices. They tested using the DBpedia-Person dataset. They compared their link prediction performance against the original RESCAL method and its non-negative variant NN-RES and showed that their method achieves higher AUC scores and diagonal confusion matrix scores.

The work of (Yang and Hospedales, 2017) presents a multi-task learning (MTL) representation learning using tensor factorisation (Tucker and TT) as a generalisation of the matrix factorisation (such as PCA) to share knowledge across tasks in fully connected and convolutional DNN layers. They compare their method to Single Task Learning (STL) vs MTL, using user-defined representation vs the learned representation on shallow and deep layer networks. The increased accuracy of learning the representation using tensor factorisation on deep layers is due to the end-to-end training of both the classifier and feature extractor.

There is also a generative model using Bayesian Tucker decomposition that is proposed by (Castellana and Bacciu, 2019). It is suitable for tree-structured data. The Markov model is an expressive model that grows in size and becomes intractable for practical problems. Tensor factorisation enables the model to be a non-parametric Bayesian model as well.

The authors of (Lacroix et al., 2020) applied CP decomposition to predict dynamic knowledge graph links. They created a 4-mode tensor of subject, predicate, object, and time. They proposed a new dataset for temporal knowledge graphs parsing Wikipedia. They compared their model to other models on their proposed dataset and other datasets to show that their results are promising.

**Image Processing:**

Eigenfaces is an algorithm for face recognition, modelling several images for the same person as an unfolded vector of each image that is stacked in a matrix, using XY matrices for P persons. Then, Eigen decomposition using principal component analysis (PCA) is applied to reduce the dimensionality of the images to use only the uncorrelated variables. The result is the eigenface with the smallest Euclidian distance to which the person resembles the most (Turk and Pentland, 1991). The authors of (M. Alex O. Vasilescu and Terzopoulos, 2002) pioneered the use of Tucker decompositions in computer vision to disentangle the multiple factors an image is composed of, such as scene structure, different facial geometries (people), expressions, head poses, lighting conditions, and imaging. They contributed TensorFaces, representing the tensor decompositions' cores as a set of facial components. They applied a HOSVD on facial images (512 × 352 decimated by a factor of 3 and cropped, yielding 7943 pixels) dataset of 28 people x 5 poses x 3 illumination conditions x 3 facial expressions x 7943 pixels. This created a 5-mode tensor $D^{28,5,3,3,7943}$, that is decomposed by HOSVD to: $D = Z \times_1 U_{people} \times_2 U_{views} \times_3 U_{illumnation} \times_4 U_{expression} \times_5 U_{pixels}$, such that $U_{people} \in R^{28 \times 28}$ spans the space of people's image parameters, $U_{views} \in R^{5 \times 5}$ spans the space of viewpoint parameters and so forth for the remaining factor matrices. Each Factor matrix is computed as $U_n = D_{(n)} V_n \Sigma^+$, where $D_{(n)}$ is mode n flattening of $D$, and computing SVD on $D_{(n)}$ to have the right matrix $V$ and the singular values in the diagonal matrix $\Sigma$, considering $U_n$ to be the left matrix of the SVD. This method generalises the eigenfaces method as a factor matrix $U_{pixels}$ is the eigenimage. Solving for the core matrix $Z$ calculates the interactions of all modes considered in this experiment, as explained in the Tucker Decomposition in section 2.1.1.2.

(Vasilescu, 2002) has also applied the Tucker decomposition to human motion as a composite of multiple actions. The author had multiple aims: to extract a human movement signature as a subset of actions (analysis), to resynthesise new motions from the learned ones (synthesis), and to recognise a specific person or action (recognition). The author defined a tensor $D^{N,M,T}$, where N is the number of people, M is the number of action classes, and T is the number of joint angle time samples. The author applied the same N-mode SVD applied in TensorFaces, $D = Z \times_1 P \times_2 A \times_3 J$. The people core matrix P has N person-specific rows containing the human motion signatures. The action core matrix A has M action-specific rows encoding action invariances across people. The last joint angle matrix J has T joint angle rows, the eigenmotions typically computed by PCA. The analysis is achieved by a change of basis to capture a person-associated motion by $C = Z \times_1 P \times_3 J$, and the change of basis to capture an action-associated motion is achieved by $B = Z \times_2 A \times_3 J$. The synthesis is achieved by the knowledge of the core tensor $Z$ capturing all multi-way interactions, core matrix A generalising the actions, and core matrix J generalising the joint angles. To synthesise an action of a new person not seen before is a Tensor completion or a regression problem to predict $D_{p,a}$ of a new person p doing action a, as $D_{p,a} = B_a \times_1 p^T$, where $B_a = Z \times_2 a_a^T \times_3 J$ for the specific action a. If the aim is to synthesise a motion for a new individual, then $p^T = d_a^T B_a^{-1}$, where $d_a^T$ is the flattened tensor in the people mode, and choosing the specific action Transpose, and the complete set of motions for the new individual is $D_p = B \times_1 p^T$. If the aim is to synthesise a new action for a known person whom we have other actions recorded for, the new action $a^T = d_p^T C_p^{-1}$, then synthesising this action for all people in the database is $D_a = C \times_2 a^T$. This makes this tensor decomposition a generative model. the recognition task is achieved in this tensor decomposition by identifying a person from action parameters as the projection $p = B_a^{-T} d$ . Similarly, identifying a person's specific action is the projection $a = C_p^{-T} d$. In both cases, the nearest neighbour algorithm returns the nearest person or action in the learned motion data d.

The previous multi-way analysis case studies were either using the tensor decomposition technique for tensor completion and prediction, or as a pre-processing step for a traditional ML learning algorithm such as clustering. The data representation is done manually. Since the advances in ANN from 2006 onwards, ANN implicitly learn the representation. For example, traditionally images are converted using Fourier Transforms to the wave domain (change of basis), now, Deeper CNNs identify the different representation hierarchy based on the complexity of the images scene contents. This process studies the structural data interactions, forming an embedding representing the data that can be used for the given machine learning task at the output layer. The depth of the neural networks adds more parameters to estimate, causing the curse of dimensionality in big data analytics. It was observed that the weights within a layer in CNN can be estimated by a 5% subset of its parameters, indicating the DL models are over-parameterised (Denil et al., 2014).

**Deep Learning:**

There are three common network compression approaches that are useful for regularisation to prevent the network from overfitting. The first is adding pooling layers that are useful for regularisation as well. These layers apply some approximations through the layers by neglecting irrelevant contents by applying different pooling functions. Another approach is adding drop-out layers that randomly set some neurons to zero; this approach is called pruning. These are considered blind compression since they do this collectively on all neurons or randomly. Other methods work to learn which neurons drop out in a structured approach, such as (Fan, Grave and Joulin, 2019) and (Knodt, 2022).

The third common ANN compression approach is Vector quantisation, which can be applied to the CNN parameters and storage requirements. These methods include binarisation (1-bit quantisation) by turning off neurons with negative values. Another Quantisation method uses lower precision, such as converting floating point types to integer types, compressing the network four times and speeding it up 2:4 times. Moreover, scalar quantisation uses k-means to cluster the weights and use representative neurons of each cluster. Also, Product quantisation divides the weights vector space into many disjoint subspaces and quantises them by raising them to different powers and, applying k-means on all and storing their cluster indices only. Finally, the residual quantisation performs clustering by k-means and then identifies the residuals to reapply the clustering on them (Gong *et al.*, 2014). Other ANN compression approaches in the literature include HashedNets (Chen *et al.*, 2015), which uses a hash function to group connection weights in hash buckets, and Layer Fusion (Graph Optimization) is another compression approach used by NVIDIA TensorRT compiler and ONNX Runtime cross-platform compiler.

Various tensor factorisation algorithms have been applied to achieve NN compression. This approach requires tensorizing and decomposing the weight matrices into a series of low-rank tensors to reduce redundant weights by using sparse representations. The authors of (Novikov *et al.*, 2015) replaced fully connected layers in CNN with a Tensor Train (TT-Layer) that they called TensorNet, which is compatible with the same training algorithm. TT Format is more immune to dimensionality curse and simpler for basic operations (addition and multiplication by a constant, summation and entry-wise tensor products, the sum of all elements and Frobenius norm) than Tucker and Hierarchical Tucker tensor decompositions. They proposed a mapping function between the weights metric and the TT-format and introduced the NN layer with weights stored in TT-format to be a TT-layer, which, when used in any NN, makes it TensorNet. A fully connected layer computes $y = Wx + b$, while a TT-layer converts all to tensors in TT-format $Y(i_1, ..., i_d) = G_1(i_1, j_1) ... G_d(i_d, j_d)X(j_1, ..., j_d) + B(i_1, ..., i_d)$, where $G_i$ are the d core tensors of the TT-format of the original weights matrix, and $Y, X$ and $B$ are the d-dimensional tensors formed from the corresponding vectors y (dependent/target/outcome variable), x (independent/predictor/feature variable(s)), and b (bias), respectively. The paper details how the loss function can be performed in the TT-Format. They achieved 200,000 times fewer parameters and compressed the size of the whole network by a factor of 7 in two-TT-Layer TensorNet compared to a two-layer fully connected network on the MNIST, and CIFAR-10 datasets, without compromising the accuracy on TT-ranks all equal to 8.

The work in (Calvi *et al.*, 2020) introduced the Tucker Tensor Layer (TTL) as an alternative to the dense weight matrices of neural networks. They also showed how the number of parameters in the neural layer is reduced while deriving a Forward and back-propagation on tensors algorithm that preserves the physical interpretability of Tucker decomposition and provides an insight into the learning process of the layer. They achieved a 66.63% compression with 82.3% accuracy compared to the 86.3% accuracy of the uncompressed model.

The Hierarchical Tucker (HT) tensor decomposition method performs better in compressing the weight matrices in Fully Convolutional layers because HT prefers the tensor with balanced dimensions lengths, as shown in (Gabor and Zdunek, 2023). The authors experimented with medium-scale CNNs on the CIFAR-10 dataset and large-scale CNNs, such as VGG-16 and ResNet-50, on the ImageNet dataset. They compared HT-2 to other tensor factorisation and other NN compression approaches to show its competitiveness in the achieved compression without much drop in accuracy. A hybrid tensor decomposition combining TT and HT is proposed by (Wu *et al.*, 2020). They compared HT formats to TT-LSTM (Yang, Krompass and Tresp, 2017) and TR-LSTM (Pan *et al.*, 2019) to show that RNNs/LSTMs in the HT format have higher compression than those in the TT format when compressing weight matrices but with worse accuracy than regular uncompressed RNNs, and that the TT format is more suitable for CNNs

Earlier NLP work used bag of words such as in (Socher *et al.*, 2013) to build a Recursive Neural Tensor Network (RNTN) that uses a high-order neural network for structured data that leverages a full 3-way tensor for aggregating children's information in binary parse trees within a natural language processing application. The authors also explained the back-propagation algorithm for tensors and used AdaGrad for this non-convex optimisation. The authors in (Weber, Balasubramanian and Chambers, 2017) presented a natural language understanding application where tensors are used to capture multiplicative interactions combining predicate, object and subject, generating aggregated representations for event prediction tasks. The Predicate Tensor approach was better in one case predicting words (which was always more accurate than predicting events), using Hard similarity scores as a percentage of cases where the similar pair had higher cosine similarity than the dissimilar pair.

The latest NLP advances are due to the better encoding algorithms such as Word2Vec, GloVe and the Transformer based models. The Transformers estimate a large number of parameters and can benefit from compression techniques such as parameter sharing across layers and low-rank approximations. These can be achieved by tensor decompositions methods such as Block-Term Tensor Decomposition (BTD), which is proposed by the authors of (Ma *et al.*, 2019). BTD combines both CP decomposition and Tucker decomposition, such that a tensor is decomposed into P Tucker decomposition, each with its core tensor and d factor matrices, such that P is the CP rank. The authors first used Single-block attention based on the Tucker decomposition to use a linear function of a set of vectors. Then they built the multi-head attention using the BTD, enabling parameter sharing across multiple blocks, higher compression (8 times fewer parameters), and lower complexity. They tested using Penn Tree Bank (PTB), WikiText-103 and One-billion language modelling tasks, and English-German neural machine translation WMT-2016 to show that their method is more compressed and more accurate than Transformer, Transformer XL, TT-format tensor factorised Transformer model, and other models using RNN, LSTM, and others.

Another generative DNN architecture model is the Restricted Boltzmann Machines (RBM) which estimates the probability distribution of various datasets. Mapping an RBM to the Tensor Networks States (TNS) has been successfully applied by (Chen *et al.*, 2018). TNS has been applied to various problems in quantum-many-body physics. The physics communities refer to Tensor Chain (TC) decomposition as the Matrix Product State (MPS), which is a special case of the Hierarchical Tucker (HT) decomposition and the simplest TNS and equivalent to the TT format. They discussed other TNS models, adding more deep layers, and how the number of parameters does not increase while the model performance increases.

**Multi-Model and Graph Data:**

In multi-modal ANN models, a simple approach is concatenating the vectors or applying an element-wise sum or product between the different modalities. This will not capture complex interactions between the different modalities. Outer-Product methods are used to capture bilinear interactions between all elements of two vectors, such as an outer product $q \otimes v$ between visual v and textual q embeddings. This will generate a massive number of parameters to estimate. Multi-modal Compact Bilinear pooling (MCB) uses FFT to compress further the outer product (Fukui *et al.*, 2016). The authors in (Ben-younes *et al.*, 2017) address the Visual Question Answering (VQA) task by using tensors to fuse visual and textual representations. They proposed a multi-modal tensor-based Tucker decomposition to capture the interactions between images and textual modalities with fewer parameters (compression) than other bilinear models. They compared the performance with other state-of-the-art models to show performance improvements.

The work in (Li *et al.*, 2020) created a multi-modal sentiment analysis (MSA) using the MOSI/CMU-MOSI dataset of the form (A, V, L), where A = $\{A_1, \ldots, A_T\}$, V = $\{V_1, \ldots, V_T\}$ and L = $\{L_1, \ldots, L_T\}$, denote the time series of the length T w.r.t. the acoustic, visual and language data, respectively. They proposed Time Product Fusion Network (TPFN) that builds on the temporal tensor fusion network (T2FN). TPFN applies implicit outer product methods across sliding time windows to capture the model interaction across modalities in the data fusion phase. CP is the method for low-rank decomposition, and regularisation on the low-rank representation handles incomplete datasets.

In (Hou *et al.*, 2019), the authors addressed the Multi-modal sentiment analysis (MSA) problem by proposing a High-order polynomial tensor pooling (PTP). PTP concatenated features form a Tensor by tensor product operation of order P to represent all possible polynomial expansions up to order P. As P increases, so does the number of parameters to learn, but the higher polynomial interactions between tensors can be captured. Using CP decomposition, the weights' tensor is compressed. Then, a Hierarchical polynomial fusion network (HPFN) is formed, assuming a 2D feature map time series. HPFN recursively learn the local temporal modalities pattern by arranging PTP in multiple layers. This borrows many features from CNN, including receptive fields, sharing parameters, scanning window, and PTP 'fusion filters'.

Graph Neural Networks (GNN) present a class of models that takes graph or network data structures as input, and it has been applied in DNN and using Tensor decomposition approaches. GNN is an active research topic and is almost reaching maturity, as presented by (Liu and Zhou, 2020a). The authors (Kwon and Chung, 2022) proposed a recursive tensor decomposition method that is based on the CP decomposition by choosing orthogonal vectors in the SVD step, creating a decomposition tree.

The work in (Hamdi and Angryk, 2019) presents tensor decomposition-based node embedding algorithms that learn node features from arbitrary types of graphs: undirected, directed, and/or weighted, without relying on computationally expensive eigen decomposition or requiring tuning of the word embedding-based hyperparameters as a result of representing the graph as a node sequence similar to the sentences in a document.

The work in (Jermyn, 2019) presents tensor trees as efficient tensor computer representations based on both optimal brute force and greedy algorithm heuristic that performs well for higher-rank tensors tree decompositions. For graph transformation, graph-tensors proposed in (Malik *et al.*, 2019) learn embeddings of time-varying graphs based on a tensor framework. There are also matrix networks proposed in (Sun *et al.*, 2018) and graph tensor neural networks (Liu and Zhou, 2020b).

## 4.3 Python Packages

Various Python packages implement multi-way Tensors and the algorithms based on them. Table 4 summarises the functions implemented by some famous packages in the literature.

*Table 4: Python Packages implementing Tensor methods*

| Package Name | Functionality |
| --- | --- |
| Tensorly (Kossaifi *et al.*, 2019) | Various Tensor decomposition, such as CP and Tucker, tensor regression, a Tensor Regression Layer (TRL), FC layer using TT format |
| scikit-tt(Gelß, 2022) | Various Tensor decomposition, TT decomposition, tensor regression |
| HOTTBOX(Kisil *et al.*, 2021) | Various Tensor decomposition |
| scikit-tensor(Nickel, 2013) | Tensor decomposition such as  INdividual Differences in SCALing (INDSCAL) CP, Tucker, DEDICOM, and RESCAL |
| ttrecipes(Ballester-Ripoll and Paredes, 2022) | Tensor regression |
| T3F(Novikov *et al.*, 2020) | Tensor Completion, Tensor Train decomposition for neural networks (NN) |
| TT_RNN(Yang, Krompass and Tresp, 2017) | A Tensorial RNN, FC, Simple RNN, LSTM and GRU using PyTorch |
| TensorNet-TF (Garipov et al., no date) | Tensorised FC layer and CNN layer |
| Spektral(Grattarola and Alippi, 2020) | Different GNN models |
| Deep Graph Library (DGL) (Wang et al., 2019) | Different GNN models |
| PyG (Fey and Lenssen, 2019) | PyTorch geometric is another GNN framework |
| TensorD (Hao et al., 2018) | Python tensor library built on Tensorflow with basic tensor operations and decompositions supporting parallel computation (e.g. GPU). |
| Tednet  (Pan, Wang and Xu, 2022) | Various neural network layer types compressed using different tensor decompositions, such as compressing an RNN layer using TR decomposition (TR_RNN). They support ResNet Layers, LSTM Layers, CNN, and Linear Layers, among others |
| Tensortools (Williams et al., 2018). | Time-shifted CP decomposition |

## 5. Conclusion and Future Directions

In this paper, we have conducted an extensive survey of multi-way analysis approaches, contrasting them with traditional linear algebra-based machine learning algorithms that rely on matrix-form pair-wise relationships. The provided tensorization unveils the expressive power of multidimensional data for enhanced multiway analysis and its integration with deep neural networks.  Building on the tutorial-style introduction presented by (Helal, 2023), we provided a comprehensive overview of tensor computing fundamentals. The accompanying Python code implementation served as a valuable resource for understanding the core algorithms, though it did not address advanced considerations such as vectorisation, parallelisation, and handling exceptional cases.

One key finding from our experiments is the pressing need to generalise machine learning packages to accommodate datasets with varying dimensions, as dictated by specific application requirements. Currently, most ML frameworks are built around traditional matrix representations or 2D/3D tensors in the context of convolutional neural networks (CNNs), treating batches as an additional dimension when necessary. As such, existing multi-way analysis implementations suffer from limitations in this regard. Another key finding is the lack of standardisation for sparse tensors and the suitable adaptation required from ML and DL algorithms stacks.

We explored various tensorization techniques, including slow-paced tensorization by adding coordinates, choosing basis functions, segmentation, and dataset merging, and surveyed both deterministic and statistical methods. Our examination of applications across diverse domains elaborated further in (Helal, 2023) underscores that multi-modal deep neural networks (DNNs) and graph neural networks (GNNs) emerge as the primary beneficiaries of tensor-based approaches. The ongoing group theoretic research outcomes and its promising direction of generalising machine learning algorithms over various mathematical structures and algebras motivate further research and development of stacks of optimised implementations.

The benefits of adopting multi-way analysis techniques are manifold: they facilitate enhanced representation of complex relationships, feature extraction from tensorized data, efficient data compression (parameter reduction), accurate prediction and classification, improved generalisation and robustness, interpretability and explainability, as well as scalability through efficient memory and processing utilisation. However, these advantages come with their own set of challenges. Data preprocessing to conform to the required tensor form remains a formidable task. Moreover, the lack of standardisation in tensor data structures across Python packages poses an obstacle to seamless integration.

Historically, Basic Linear Algebra Subprograms (BLAS) libraries have played a vital role in optimising numerical recipes for solving equations and various linear algebra computations. BLAS started with vector operations only, then matrix operations, then matrix-matrix operations. Recently, Tensor-tensor operations BLAS for the fourth-order tensors were implemented (Liu and Wang, 2017), including tensor (Kronecker) product, KhatriRao product, Hadamard product, tensor contraction, t-product, or L-product. A similar optimisation level is needed for tractable tensor-tensor operations on variable-order tensors. This also needs to be interoperable between different deep-learning frameworks and parallel hardware platforms. Tensorised NN Frameworks can be built with all Tensorised Layer types, tensorised activation functions, and tensorised forward and backward propagation algorithms such as the SGD with DMRG algorithms, AutoDiff (Paszke et al., 2017) and DDSP (differentiable digital signal processing) (Engel et al., 2020)

Looking ahead, the development of tensorized neural network frameworks should encompass a wide range of tensorized layer types, activation functions, and forward/backward propagation algorithms. Section 4 of this paper illustrated various case studies on tensorizing neural networks at different stages of the architecture, but ongoing research is required to determine the optimal number and type of tensorizations and their collective performance impact. The choice of tensor decomposition techniques, such as CP, Tucker, HT, TT, or others, should be further investigated to strike a balance between compression and performance enhancement, as well as interpretability across different models and applications. Establishing standardised benchmarks with clear metrics will facilitate the comparison of future proposals.

Lastly, the generalisation of graph neural networks (GNNs) and dynamic graph networks (DGNs) to handle datasets with unconstrained topologies, such as hypergraphs with hierarchical connections, represents a promising research avenue. Systematic evaluations of expressiveness, achieved through tensorization, can elucidate the performance of newly proposed models (Errica, Bacciu and Micheli, 2020). Additionally, exploring time-evolving graphs, temporal-spatial learning, or online learning using tensorized models on graphs promises exciting directions for enhancing and evaluating tensorization's impact on compression, expressiveness, and accuracy trade-offs.

In summary, this paper comprehensively explores tensorization and its potential to revolutionise deep learning models and multi-way analysis. Addressing the challenges and further investigating these methodologies will pave the way for more efficient, expressive, and adaptable machine learning systems as we move forward.